# How the Brain might use Division

Kieran Greer, Distributed Computing Systems, Belfast, UK.
http://distributedcomputingsystems.co.uk
Version 1.1

**Abstract** – One of the most fundamental questions in Biology or Artificial Intelligence is how the human brain performs mathematical functions. How does a neural architecture that may organise itself mostly through statistics, know what to do? One possibility is to extract the problem to something more abstract. This becomes clear when thinking about how the brain handles large numbers, for example to the power of something, when simply summing to an answer is not feasible. In this paper, the author suggests that the maths question can be answered more easily if the problem is changed into one of symbol manipulation and not just number counting. If symbols can be compared and manipulated, maybe without understanding completely what they are, then the mathematical operations become relative and some of them might even be rote learned. The proposed system may also be suggested as an alternative to the traditional computer binary system. Any of the actual maths still breaks down into binary operations, while a more symbolic level above that can manipulate the numbers and reduce the problem size, thus making the binary operations simpler. An interesting result of looking at this is the possibility of a new fractal equation resulting from division, that can be used as a measure of good fit and would help the brain decide how to solve something through self-replacement and a comparison with this good fit.

## 1   Introduction

One of the most fundamental questions in Biology or Artificial Intelligence is how the human brain performs mathematical functions. How does a neural architecture that may organise itself mostly through statistics, know what to do? There is increasing understanding about the analogue properties of neurons [1][13][14]. That is, they can send a variety of signals and not just the traditional on-off model. While that is the case, computers still use the binary system because it is reliable and more easily understood. Therefore, when solving the mathematical problem, the binary system still has to be considered as the main player and so the decimal system, for example, has to be translated over to it. How a neural system would do this is still not very well understood. Then when it goes to processing the numbers, the system would need to understand quite complex operations that are exact and iterative. One possibility therefore, is to extract the problem to something more abstract. This becomes clear when thinking about how the brain handles large numbers, for example to the power of something, when simply summing to an answer is not feasible. A computer can perform this maths with logic gates and so the question is, can the brain components do the same thing? Missing from this argument is the fact that the neurons are expected to perform in a completely blind manner, when in fact, the human body provides senses and a nervous system to give some type of guidance. Therefore, to start with, images of the number system can give some sort of base to work from.

In this paper, the author suggests that the maths question can be answered more easily if the problem is changed into one of symbol manipulation and not just number counting. If symbols can be compared and manipulated, maybe without understanding completely what they are, then the mathematical operations become relative and some of them might even be rote learned. To anchor the solution therefore, the brain can use images of the number system that would be learned before the maths itself. For the decimal number system, for example, the brain learns the symbols '0' to '9' first. Everything after that is the manipulation of these symbols. The system to be proposed in this paper may also be a suggestion for an alternative to the traditional computer binary system. Any of the actual maths still breaks down into binary operations, while a more symbolic level above that can manipulate the numbers and reduce the problem size, thus making the binary operations smaller in magnitude. Essentially, the higher level breaks the problem down into smaller chunks that is understood to certain orders of magnitude. The binary maths is then performed only on what is left. This paper therefore proposes a theory for how mathematics can be carried out through the manipulation of grids and/or arrays of these symbols, in a system that may be closer to what the human brain uses. While the theory has arisen by thinking about the human brain, it will simply be called a system in the rest of the paper and a reference will not be made to the human brain each time. An interesting result of looking at this is the possibility of a new fractal equation resulting from division, that can be used as a measure of good fit and would help the brain decide how to solve something through self-replacement and a comparison with this good fit.

The rest of the paper is organised as follows: Section 2 describes the theory of the mathematical process. Section 3 gives an algorithm for each mathematical operation – add, subtract, multiply and divide, in turn. Section 4 discusses a new equation of good fit that may be a fractal and also results from the number system. Section 5 describes some related work, while section 6 gives some final conclusions.

## 2   Theory

The theory starts with the assertion that all numbers are combinations of the base number set. To solve the problem therefore, you need to break it down into problems over this base number set that are either very small or easier to process. Therefore, instead of solving a small set of large number problems, it gets broken down into a large set of small number problems with some automatic transpositions. These smaller problems map more closely to the symbol set that the system understands, even to the point where some small operations are simply rote learned. The binary array is replaced by a grid or table format for this solution, with each column representing a different decimal digit. A second higher level then converts what would be iterative counts into more singular transposition operations that require some understanding at the symbolic level.

### 2.1   Bitwise Grid

The first part of the system is the bitwise grid that represents the number system itself, shown in Figure 1. The grid is the size of the number set horizontally, and then vertically it represents orders of magnitude over the base value. For the decimal number system therefore, the unit numbers 0 to 9 are at the bottom level, then the 10's from 10 to 90 at the second level, the 100's are at the third level and so on, to as far as is required. The mathematical process then requires that numbers are broken down into whole base parts and remainders, where whole base parts can be subject to transpositions and most of the actual maths takes place over the remainder parts. Breaking the numbers down means that the process is now distributed and different areas of the table would be used for the single operation. The number parts may also be indexed so that they can be consistently updated and re-joined at the end. The transpositions can require a number to move to a neighbouring cell in any of the 4 directions, or moving up/down a level if it moves off the right/left edges of the table. After each phase, cells on the same level are added together and if any phase produces new numbers, then they can be similarly broken down and the process repeated. After the maths is completed, the result is then to re-join these indexed parts. This is the symbolic replacement of zeros in the result by non-zeros from any other part of the number. When using this table, rows are counted from the bottom-up and columns from left to right. So, for example, Units would be in row 1 and cell (row, column) [3, 5] represents the number 4000.

| *OM* | 1 | 2 | 3 | 4 | 5 | 6 | 7 | 8 | 9 | 10 |
|---|---|---|---|---|---|---|---|---|---|---|
| 7 | 0 | 1 | 2 | 3 | 4 | 5 | 6 | 7 | 8 | 9 |
| 6 | 0 | 1 | 2 | 3 | 4 | 5 | 6 | 7 | 8 | 9 |
| 5 | 0 | 1 | 2 | 3 | 4 | 5 | 6 | 7 | 8 | 9 |
| 4 | 0 | 1 | 2 | 3 | 4 | 5 | 6 | 7 | 8 | 9 |
| 3 | 0 | 1 | 2 | 3 | 4 | 5 | 6 | 7 | 8 | 9 |
| 2 | 0 | 1 | 2 | 3 | 4 | 5 | 6 | 7 | 8 | 9 |
| 1 | 0 | 1 | 2 | 3 | 4 | 5 | 6 | 7 | 8 | 9 |
| Units | 0 | 1 | 2 | 3 | 4 | 5 | 6 | 7 | 8 | 9 |

Figure 1. Bitwise Grid. Moving vertically changes the order by the number system base, horizontally changes the number by a unit at that level.

### 2.2    Separate a Number into Parts

A lot of the usefulness of this system is the ability to recognise the orders of magnitude and therefore remove that from the calculation. The maths takes place over sets of smaller numbers only. To do this, any larger number needs to be split-up into parts, which represent each order of magnitude that it has. This is very easily done simply by creating a new number for every non-zero digit and replacing any removed digits with 0. For example, the number 12045 would produce the number parts: 10000, 2000, 40 and 5. This should be done both for the number being operated on and the operand.

### 2.3 Orders of Magnitude

The system only performs maths over small whole numbers, apart from the division operation. If the number is large, then it is represented by a higher level in the table. For example, if the number is 10000, then in Figure 1, it is placed in the cell at row 5 and column 2. Division and subtraction can lead to negative results and so each number, represented by a cell in the grid, can store a negative sign as well. In division, for example, the positive order of magnitude may relate to the size of what the whole integer part would be and the negative order of magnitude would relate to how many orders the fractional part needs to be moved to be that whole number. For example, if the number is 10000.2, then it would be assigned to the cells [5, 2] (instead of [4, 2]) and also [0, 2] (for the 2 units). A negative order of magnitude of 1 is then stored with the first (highest order) cell. Maths would be carried out on this as normal, but at the end, the negative order of magnitude will move the decimal point to the left, that number of places, again.

### 2.4 Base Number Transpositions

The second part of the system is to transpose the number using the table of Figure 1. A number is changed by a unit or order of magnitude of the number base, depending on the mathematical operator.

- To transpose by an order of magnitude requires that it is wholly divisible by the base number, in this case, wholly divisible by 10.
- To transpose by a unit requires that the numbers are at the same level, or on the same row in the table.
- The maths for any transposition makes use of the non-zero digits, represented by the cell value and always located on the left-hand side of any number.

Then the transposition is automatic. Considering the mathematical operators of add, subtract, multiply and divide therefore, the table of Figure 1 would be used as follows:

- **Add:** to add a unit number at the same level of the table row, move the number to the right. For example, adding 3 to 5, moves the number to cell value 8 at the Units level. Adding 100 to 200, moves the number to cell value 3 at the 100's level.
- **Subtract:** similarly, to subtract a unit number at the same level of the table row, move the number to the left.
- **Multiply:** to multiply by the base number, move the number vertically upwards in the table. So, to multiply 10 by 100, moves the 10 cell up two levels in the table, as indicated by 2 zeros in the multiplication number.
- **Divide:** similarly, to divide by the base number, move it down levels in the table.

As the mathematics takes place at the left-hand side of the number, a note of the number of zeros can be made, for example, and then any sum or product result can replace the LHS non-zero part, while the zero digits remain the same. The level therefore defines the number of zeros and the maths takes place over the non-zero digits only. The system is ideally setup for whole number operations in the positive integer range, but fractions and negative numbers have to be accommodated. This would typically require manipulation using negative orders of magnitude instead of positive ones.

## 3    Mathematical Operators

This section gives the algorithms that each of the four mathematical operations might use. They are not definitive and could be changed, so this is only an example. The maths occurs across each table row separately, with the most complicated operations moving cells off the side of a table to a new row above or below the current one. Division is the only problem and it can use larger numbers. The actual maths is then switched over to a binary and traditional format, with the result being switched back again. The numbers in each row are then re-joined together to give a final total. If the process leads to intermediate results, then those numbers can replace the original set and be split again, before repeating the process. When re-joining and summing, it is simply a matter of placing the non-zero digits from each result part in the final number at the appropriate place. Appendix A traces through some examples of using these algorithms.

### 3.1   Sums Over Numbers with Negative Orders of Magnitude

If doing sums over numbers with negative orders of magnitude, then the orders need to be made the same first, which would be the largest negative order value. Any numbers with a smaller order are filled out with zeros on their right-hand side. For example, consider the sum 0.7 + 0.05. This would be represented in the system as 7 (-1) and 5 (-2), where the brackets are the negative orders of magnitude, but these numbers are currently incompatible and would need to be re-written as 70 (-2) + 5 (-2) = 75 (-2) = 0.75.

### 3.2   Addition

With addition, the sum can result in numbers larger than the row level the maths is performed on. When this happens, the number needs to be split again into a wholly divisible part for the next level up and one for the current level. These two levels then need to be re-calculated with the split number and any other number at the level above. The following algorithm can perform the addition operation.

1) Split the number $X_i$ and place the parts at each level in the table.
2) Split the addend $A_j$ and place the parts at each level in the table.
3) While any table level (row) has more than 1 cell entry:
   a) From the lowest order of magnitude to the highest, find the next table level L with more than 1 cell entry:
      i) Add the digits of cells with entries at level L together.
      ii) If the new number has more than 1 digit, then it can be split again.
         (1) The split produces new numbers for 2 table levels – the current level L and the level immediately above (L+1), and so these levels need to be re-calculated.
         (2) Place the numbers for those levels into the appropriate cells again.
4) Re-join all of the cell numbers to create the number result.

### 3.3   Subtraction

With subtraction, the sum can result in negative numbers and this is usually managed by borrowing a digit from the next order above. With the table this is quite easily realised by changing two cell numbers and adding new entries in some other cells. When this happens, there may need to be more calculations over the new cell entries, but it is largely a visual process without too much additional calculation. Subtraction can also give a final result that is negative, or an intermediate result that requires a positive number to be added to a larger negative one. When this happens, it is easier to switch the signs of the level cells involved, for the calculation, and then switch the result sign back again. To keep track of the signs the number parts may need to be indexed. It is noted that these algorithms may appear to be quite complicated and traditional binary mathematics can do this operation simply by reversing the sign of the subtrahend and then adding the two numbers. While this algorithm is more involved, the question would be if it can still be mostly automated, as each individual step is relatively simplistic. The following algorithm can perform the subtraction operation.

1) Split the minuend $X_i$ and place the parts at each level in the table.
2) Split the subtrahend $S_j$ and place the parts at each level in the table.
3) While any table level has more than 1 cell entry:
   a) From the highest order of magnitude to the lowest, find the next table level L with more than 1 cell entry:
      i) Check if the minuend cell digit $C_{xil}$ is smaller than the subtrahend cell digit $C_{sjl}$.
      ii) If the cell digit is smaller, a digit needs to be borrowed from a higher level.
         (1) If a higher level has a minuend cell:
            (a) Move to the next level with a minuend cell $C_{xil2}$.
            (b) Add the digit 1 in-front of the current cell digit $C_{xil}$ and move the minuend cell $C_{xil2}$ one place to the left.
            (c) For each level between the two levels L and $L_2$, add a new entry in the '9' digit cell.
            (d) This should in fact be the only minuend entry at those levels, but the subtrahend may also have an entry.

        (2) If a higher level does not have a minuend cell, then store the negative calculation as the result.

   iii) For the current level L, subtract the subtrahend cell $C_{sjl}$ digit from the minuend cell $C_{xil}$ digit(s) and store the result. The result will be a single digit.

4) While there are both negative and positive signed digits:
   a) From the highest order of magnitude to the lowest, find the next two consecutive table levels with a cell sign change.
   b) This should be a higher level $L_2$ with the negative sign and a lower level L with the positive sign.
   c) Index these cell numbers and reverse their signs.
   d) Perform a subtraction over these two levels only. That will require a borrow operation to create a maximum valued minuend at level L, with a negative sign, and move the existing digits one cell left.
   e) The subtraction then takes place over the new minuend value and the subtrahend value at level L.
   f) Switch the cell signs back again. This should make them both negative.
5) Re-join all of the cell numbers to create the number result.

### 3.4   Multiplication

Multiplication uses a bitwise operation of moving the number cells up levels depending on the order of magnitude of the multiplier. It is then required to perform a product sum of the non-zero digits in the multiplicand and multiplier, and replace the non-zero digits in the result part with these.

1) Split the multiplier M into parts.
2) From the highest order of magnitude to the lowest for each multiplier part $M_j$:
   a) Split the multiplicand X into parts.
   b) From the highest order of magnitude to the lowest for each multiplicand part $X_i$:
   c) Place multiplier part $M_j$ in the appropriate table cell $C_{mj}$.
   d) Place multiplicand part $X_i$ in the appropriate table cell $C_{xi}$.

- i) Move the multiplicand cell $C_{xi}$ vertically upwards, relative to the order of magnitude of the multiplier cell $C_{mj}$.
- ii) Multiply the cell digits in the multiplicand and multiplier together and replace the multiplicand digit $C_{xi}$ by them.
- iii) If the cell number $C_{xi}$ has more than 1 digit, then it can be split again.
    - (1) The split produces new numbers for 2 table levels – the current level L and the level immediately above (L+1), and so these levels need to be re-calculated.
    - (2) If there is now more than 1 multiplicand cell at a level, then this requires an Addition operation at that level, which may move the cell digits again.
    - (3) After these operations, the process can continue to the next multiplicand level.
- e) Save the set of cell values for the multiplier part $M_j$.
3) Add all of the saved cell values for each multiplier part and perform an addition on them at each level.
4) Re-join all of the cell numbers to create the number result.

## 3.5    Division

As should be expected, division is the most difficult operation. Division uses a bitwise operation of moving the number cells down levels depending on the order of the multiplier. It is then required to perform a division sum of the non-zero digits in the dividend and divisor, and replace the non-zero digits in the result with these. The number parts are then re-joined to give the result. With division however, it is not possible to split the divisor into parts, but it is still possible to split the dividend into parts and partially reduce the number of calculations as follows:

1) While there is a dividend number to process:
   a) Split the dividend D into parts.
   b) The divisor has to be considered as a whole number, so place it in the cell with exactly that value or the next cell immediately above the value, for comparison purposes.
   c) From the highest order of magnitude to the lowest for each dividend part $D_i$:
      i) Move the dividend part $D_i$ vertically downwards, until it is at a cell immediately above the divisor cell.

        ii) Do the division of the divisor on the moved dividend part (using binary computations).

          (1) The result of the calculation is the division number plus the order of magnitude that was removed. Store both for the result.

          (2) If the division produced a remainder, then store that with its correct order of magnitude.

    d) After dividing all of the dividend parts by the divisor, add their division results together, including the orders of magnitude, for a division total.

    e) Add all of the remainder values together, including the orders of magnitude, and set this to be the new dividend value.

    f) Repeat the whole process while the divided is larger than the divisor.

2) Any value that is left is the remainder of the division.

## 4    Division as a Fractal

Division might be looked at as a measure of fit. How well does the divisor fit into the dividend? If considering it this way, then there might not be a precise numerical measurement, but judgement about a goodness of fit can be made on what could be more abstract types of object. What would this judgement be for? If there is some amount of a resource, energy or ensemble for example, then it would be helpful to know how best to use it. If it is used-up by a single entity, other entities would be left out and so it can provide a measure of constraint over how best to fit the available entities into the available resource. The division problem of this paper would work best if the divisor could also be split up into parts. Unfortunately, there are problems when trying to do that, which is especially true when thinking about prime numbers. For example, the problem of 425 / 23 = 18.48 has a prime number as the divisor. The dividend can be split into 400, 20 and 5, as described in section 3.5, but the divisor has to stay as it is. Considering splitting the divisor however has led to a type of chain equation, where the result of one part is the dividend to be measured by the next part. The chain equation does involve splitting the divisor up into parts, but there is a catch to it that is described shortly.

So firstly, what values would the divisor value of 23 be split into? In fact, it can be split into anything, so long as the parts sum up to the original total of 23. In Equation 1 for example, the divisor is split into the values 13 and 10. The following chain equation can then be used to perform the division:

1.1.    425 / 13 = 32.69
1.2.    32.69 * (10 / 23) = 14.21
1.3.    32.69 – 14.21 = 18.48

Equation 1. Chain equation for Division with 2 divisor parts.

If the divisor is split into 3 parts, say 12, 9 and 2, then Equation 2 gives the chain equation that will produce the same result:

2.1.    425 / 12 = 35.416
2.2.    35.416 * (9 / 23) = 13.858
2.3.    13.858 * (2 / 9) = 3.079
2.4.    35.416 – 13.858 – 3.079 = 18.48

Equation 2. Chain equation for Division with 3 divisor parts.

The catch in the problem is clear – the whole divisor number is required at the second division stage each time and so it cannot be got rid of completely. But because the original divisor is part of the chain equation, it would be possible to replace it again with a version of the chain equation. If you do that using exactly the same numbers, then they cancel each other out and leave only the whole divisor, but the number sequences could also be different each time and the process of replacement with something self-similar looks a bit like a fractal equation [3]. The equation in a general sense is as follows:

Dividend / $d_1$ = $r_1$

$r_1$ * $d_2$ / Divisor = $r_2$

r2 * d3 / d2 = r3

r3 * d4 / d3 = r4

…

where

d1 + d2 + d3 + d4 … = Divisor

and

r1 - r2 - r3 - r4 … = Result.

Equation 3. Chain equation for Division in a general sense.

So how might the equation be used in practice? What if the original quantity is unknown and cannot be properly measured, but divisor parts can be measured. If the chain equation is calculated and gives some result plus the original unmeasured part, then maybe the chain result can be compared with the original problem for similarity. If they are still similar, then the chain equation has not altered the original problem and it is a good fit or solution for it. This could be especially true with something like the brain, where patterns cycle in sequences and so the fit is not just once, but over many iterations.

## 5   Related Work

Symbolic computing was recognised early on as important to Artificial Intelligence, where a recent review is given in [8]. Some of the following text is taken from [7], Chapter 2: John McCarthy tried to introduce learning into a computer program, to allow it to reason like a human. His ultimate objective was to write a program that learned from experience as well as a human does. Along with Marvin Minsky they tried to design a system based on the principle that it has common sense if it can deduce for itself consequences from what it is told and what it already knows. They chose to base the system on a logic-based language called Lisp [10]. This is a high-level language of functions and nested functions over symbolic expressions. Some type of hierarchical structure in AI is assumed and it would be interesting if this paper extends that symbolic processing down into low-level binary functions. Newell and Simon [11] also supported a symbolic approach. With their 'Physical Symbol System Hypothesis' they noted that:

'Symbols lie at the root of intelligent action, which is, of course, the primary topic of Artificial Intelligence. For that matter, it is a primary question for all of Computer Science. For all information is processed by computers in the service of ends, and we measure the intelligence of a system by its ability to achieve stated ends in the face of variations, difficulties and complexities posed by the task environment.'

They therefore stated the need for a symbolic representation of the environment, so that a computer can understand it, and this requires some sort of formal specification or logic. However, a universal machine that can create, understand and use these symbols remains a problem to be solved. The earlier part of the paper has suggested how it might be converted across into computer hardware [2]. Specifically, that would be the Central Processing Unit (CPU) and the Arithmetic and Logic Unit (ALU) that performs the mathematical calculations. The binary system is used in computers because it works very well and is easier to implement. The simple 2-value system is not prone to error. Logic gates can produce these on-off values very easily and while a ternary gate with 3 states has been suggested [12], it has never been implemented because of the complexity with analogue output values. The symbolic representation of this paper however can probably be realised in a binary system and the actual maths can also be performed using the binary number system. While there is another level of complexity, it can probably be broken down into binary operations as well. The main question would be if the new level of complexity would require too many additional operations to be economical.

It is interesting that fractals [9][3] have been written about earlier [6] in terms of natural systems relating to the human brain, and the research of this paper has resulted in a similar type of argument. There are two transitions that could be measured. One is for the divisor part ($d_1$, $d_2$, … in Equation 3), moving from one value to the next and the other is for the result part ($r_1$, $r_2$, … in Equation 3) of each calculation. There is lots of evidence from earlier work that smooth transitions are more desirable in terms of energy consumption and entropy, or disruption [5][6]. The system prefers to work in a less disruptive state, but the disruption can become significant and may even help to define separate stages. It is the case that fractals in nature start with a larger value and transition smoothly to smaller and smaller variations. This would be useful for cognitive processes, where the first task would receive the largest signal but also be the most important, leading to further tasks maybe even further in time being represented by smaller signals. If there is a jump to a larger signal again, then maybe that is a milestone that should be achieved, before working out what to do from there on. The idea therefore, is that measuring the transition amounts can provide some level of guidance, and summing to the original energy total is a constraint for how sub-patterns should organise themselves. A sequence of patterns is just that – a line of patterns where the first one should be the strongest and the final one the weakest. The idea of a chain equation, but more probably a number series is also used in [4] with relation to evaluating behaviours, and therefore it is a similar but slightly different equation.

## 6   Conclusions

This paper has made suggestions as to how the brain can be helped to perform fundamental maths through the use of symbols. It is not unreasonable to include sensory input, because otherwise the brain is blind. The system would not just count numbers in a traditional sense, but manipulate symbols as well. A key result of this is that the more difficult maths is restricted to relatively small operations, while the larger calculations are replaced by symbolic transpositions. If this works for the decimal number system, then it should probably work for any number system and because it is still essentially a bitwise process, algorithms that a CPU might use, for example, have been described.

Then a new equation has been realised that looks like it could be a fractal. The equation could help with measuring an unknown quantity by comparing it with similar known quantities. It is suggested that this type of comparison can be a guidance to how the brain may link-up and fire patterns. It is obvious that the brain cannot consume more energy than is put into it, but the equation may help to establish a more measured approach as to how that energy gets used. Any set of sub-components must equate to the enclosing parent component in terms of some resource and so this is a constraint on how the sub-components can organise themselves and fire together.

## Appendix A – Some Examples

This appendix gives some worked examples of how a computer system might use the number table to perform fundamental mathematical operations.

**1. Addition Example: 55 + 150 = 205**

This is the fairly simple addition of two numbers. The numbers have been split into their orders of magnitude parts and added to the grid, shown below in Figure 2. Note that the cell [2, 6] has 2 entries, one for the number and one for the addendum. The Maths takes place over cell [2, 6]. This leads to a value of 10 for that cell.

|   |            | 1 | 2 | 3 | 4 | 5 | 6 | 7 | 8 | 9 | 10 |
|---|------------|---|---|---|---|---|---|---|---|---|----|
| 8 | 10000000's | 0 | 1 | 2 | 3 | 4 | 5 | 6 | 7 | 8 | 9  |
| 7 | 1000000's  | 0 | 1 | 2 | 3 | 4 | 5 | 6 | 7 | 8 | 9  |
| 6 | 100000's   | 0 | 1 | 2 | 3 | 4 | 5 | 6 | 7 | 8 | 9  |
| 5 | 10000's    | 0 | 1 | 2 | 3 | 4 | 5 | 6 | 7 | 8 | 9  |
| 4 | 1000's     | 0 | 1 | 2 | 3 | 4 | 5 | 6 | 7 | 8 | 9  |
| 3 | 100's      | 0 | **1** | 2 | 3 | 4 | 5 | 6 | 7 | 8 | 9  |
| 2 | 10's       | 0 | 1 | 2 | 3 | 4 | **5** | 6 | 7 | 8 | 9  |
| 1 | Units      | 0 | 1 | 2 | 3 | 4 | **5** | 6 | 7 | 8 | 9  |

Figure 2. Cells for the addition sum. Note that cell [2, 6] has 2 entries.

As the value 10 is 2 digits, it moves that cell up to the next level, to the 1 digit or cell [3, 2] position, as shown in Figure 3.

|   |            | 1 | 2 | 3 | 4 | 5 | 6 | 7 | 8 | 9 | 10 |
|---|------------|---|---|---|---|---|---|---|---|---|----|
| 8 | 10000000's | 0 | 1 | 2 | 3 | 4 | 5 | 6 | 7 | 8 | 9  |
| 7 | 1000000's  | 0 | 1 | 2 | 3 | 4 | 5 | 6 | 7 | 8 | 9  |
| 6 | 100000's   | 0 | 1 | 2 | 3 | 4 | 5 | 6 | 7 | 8 | 9  |
| 5 | 10000's    | 0 | 1 | 2 | 3 | 4 | 5 | 6 | 7 | 8 | 9  |
| 4 | 1000's     | 0 | 1 | 2 | 3 | 4 | 5 | 6 | 7 | 8 | 9  |
| 3 | 100's      | 0 | **1** | 2 | 3 | 4 | 5 | 6 | 7 | 8 | 9  |
| 2 | 10's       | 0 | 1 | 2 | 3 | 4 | 5 | 6 | 7 | 8 | 9  |

| | | | | | | | | | | |
|---|---|---|---|---|---|---|---|---|---|---|
| 1 | Units | 0 | 1 | 2 | 3 | 4 | **5** | 6 | 7 | 8 | 9 |

Figure 3. Occupied cells after adding the 50 cells together.

That leads to cell [3, 2] having 2 entries, when another addition operation is required to add them together. This leads to a value of 2 for that cell, moving it one position to the right, as shown in Figure 4.

| | | 1 | 2 | 3 | 4 | 5 | 6 | 7 | 8 | 9 | 10 |
|---|---|---|---|---|---|---|---|---|---|---|---|
| 8 | 10000000's | 0 | 1 | 2 | 3 | 4 | 5 | 6 | 7 | 8 | 9 |
| 7 | 1000000's | 0 | 1 | 2 | 3 | 4 | 5 | 6 | 7 | 8 | 9 |
| 6 | 100000's | 0 | 1 | 2 | 3 | 4 | 5 | 6 | 7 | 8 | 9 |
| 5 | 10000's | 0 | 1 | 2 | 3 | 4 | 5 | 6 | 7 | 8 | 9 |
| 4 | 1000's | 0 | 1 | 2 | 3 | 4 | 5 | 6 | 7 | 8 | 9 |
| 3 | 100's | 0 | 1 | **2** | 3 | 4 | 5 | 6 | 7 | 8 | 9 |
| 2 | 10's | 0 | 1 | 2 | 3 | 4 | 5 | 6 | 7 | 8 | 9 |
| 1 | Units | 0 | 1 | 2 | 3 | 4 | **5** | 6 | 7 | 8 | 9 |

Figure 4. Occupied cells after adding the 100 cells together.

There is no more maths to perform, so it is a matter of dropping the cell values down into their correct positions, which leads to the result of 205.

2. **Subtraction Example: 10450 – 555 = 9895**

This example subtracts 555 from 10450. The numbers have been split into their orders of magnitude parts and added to the grid, shown below in Figure 5. Note that the cell [2, 6] again has 2 entries, one for the minuend and one for the subtrahend. The maths takes place over the rows 2 and 3 in this case. Moving from the higher orders of magnitude to the lower ones, the first row to be processed is row 3, but this requires the number 5 to be subtracted from the number 4.

| | | 1 | 2 | 3 | 4 | 5 | 6 | 7 | 8 | 9 | 10 |
|---|---|---|---|---|---|---|---|---|---|---|---|
| 8 | 10000000's | 0 | 1 | 2 | 3 | 4 | 5 | 6 | 7 | 8 | 9 |
| 7 | 1000000's  | 0 | 1 | 2 | 3 | 4 | 5 | 6 | 7 | 8 | 9 |
| 6 | 100000's   | 0 | 1 | 2 | 3 | 4 | 5 | 6 | 7 | 8 | 9 |
| 5 | 10000's    | 0 | 1 | 2 | 3 | 4 | 5 | 6 | 7 | 8 | 9 |
| 4 | 1000's     | 0 | 1 | 2 | 3 | 4 | 5 | 6 | 7 | 8 | 9 |
| 3 | 100's      | 0 | 1 | 2 | 3 | 4 | 5 | 6 | 7 | 8 | 9 |
| 2 | 10's       | 0 | 1 | 2 | 3 | 4 | 5 | 6 | 7 | 8 | 9 |
| 1 | Units      | 0 | 1 | 2 | 3 | 4 | 5 | 6 | 7 | 8 | 9 |

Figure 5. Cells for the subtraction sum. Note that cell [2, 6] has 2 entries.

Because the minuend is in a lower cell, it needs to borrow a value from a higher cell. It can borrow from the level 5 cell, and then add a new entry at each 9 digit cell for the level in-between, as shown in Figure 6.

|   |            | 1 | 2 | 3 | 4 | 5  | 6 | 7 | 8 | 9 | 10 |
|---|------------|---|---|---|---|----|---|---|---|---|----|
| 8 | 10000000's | 0 | 1 | 2 | 3 | 4  | 5 | 6 | 7 | 8 | 9  |
| 7 | 1000000's  | 0 | 1 | 2 | 3 | 4  | 5 | 6 | 7 | 8 | 9  |
| 6 | 100000's   | 0 | 1 | 2 | 3 | 4  | 5 | 6 | 7 | 8 | 9  |
| 5 | 10000's    | 0 | 1 | 2 | 3 | 4  | 5 | 6 | 7 | 8 | 9  |
| 4 | 1000's     | 0 | 1 | 2 | 3 | 4  | 5 | 6 | 7 | 8 | 9  |
| 3 | 100's      | 0 | 1 | 2 | 3 | 14 | 5 | 6 | 7 | 8 | 9  |
| 2 | 10's       | 0 | 1 | 2 | 3 | 4  | 5 | 6 | 7 | 8 | 9  |
| 1 | Units      | 0 | 1 | 2 | 3 | 4  | 5 | 6 | 7 | 8 | 9  |

Figure 6. Occupied cells after borrowing a 100's unit from the 10000's level.

The maths can then be carried out at level 3, which leads to a total value of 9 there. At level 2 the maths cancels the two digits out, leading to a total value of zero, as shown in Figure 7.

|   |            | 1 | 2 | 3 | 4 | 5 | 6 | 7 | 8 | 9 | 10 |
|---|------------|---|---|---|---|---|---|---|---|---|----|
| 8 | 10000000's | 0 | 1 | 2 | 3 | 4 | 5 | 6 | 7 | 8 | 9  |
| 7 | 1000000's  | 0 | 1 | 2 | 3 | 4 | 5 | 6 | 7 | 8 | 9  |
| 6 | 100000's   | 0 | 1 | 2 | 3 | 4 | 5 | 6 | 7 | 8 | 9  |
| 5 | 10000's    | 0 | 1 | 2 | 3 | 4 | 5 | 6 | 7 | 8 | 9  |

| 4 | 1000's | 0 | 1 | 2 | 3 | 4 | 5 | 6 | 7 | 8 | 9 |
| 3 | 100's  | 0 | 1 | 2 | 3 | 4 | 5 | 6 | 7 | 8 | 9 |
| 2 | 10's   | 0 | 1 | 2 | 3 | 4 | 5 | 6 | 7 | 8 | 9 |
| 1 | Units  | 0 | 1 | 2 | 3 | 4 | 5 | 6 | 7 | 8 | 9 |

Figure 7. Occupied cells after subtractions at levels 3 and 2.

There is still a value of 5 to subtract at the units level. There is no minuend at this level and so the minuend needs to borrow a value from a higher level, which is level 3 in this case. That moves the level 3 cell one place to the left and adds a new cell at digit 9 in level 2. The units level now contains 10 – 5, in terms of cell placings it could also be worked out, but the result is the cell digit 5. This final result is shown in Figure 8.

|   |             | 1 | 2 | 3 | 4 | 5 | 6 | 7 | 8 | 9 | 10 |
|---|-------------|---|---|---|---|---|---|---|---|---|----|
| 8 | 10000000's  | 0 | 1 | 2 | 3 | 4 | 5 | 6 | 7 | 8 | 9  |
| 7 | 1000000's   | 0 | 1 | 2 | 3 | 4 | 5 | 6 | 7 | 8 | 9  |
| 6 | 100000's    | 0 | 1 | 2 | 3 | 4 | 5 | 6 | 7 | 8 | 9  |
| 5 | 10000's     | 0 | 1 | 2 | 3 | 4 | 5 | 6 | 7 | 8 | 9  |
| 4 | 1000's      | 0 | 1 | 2 | 3 | 4 | 5 | 6 | 7 | 8 | 9  |
| 3 | 100's       | 0 | 1 | 2 | 3 | 4 | 5 | 6 | 7 | 8 | 9  |
| 2 | 10's        | 0 | 1 | 2 | 3 | 4 | 5 | 6 | 7 | 8 | 9  |
| 1 | Units       | 0 | 1 | 2 | 3 | 4 | 5 | 6 | 7 | 8 | 9  |

Figure 8. Occupied cells after subtractions at levels 1.

There is no more maths to perform, so it is a matter of dropping the cell values down into their correct positions, which leads to the result of 9895.

### 3. Multiplication Example: 40 x 50 = 2000

This is a multiplication example of 40 times 50: These numbers are 4 and 5 at the ten's level, as indicated by having only 1 zero in each number. It is therefore noted that to multiply by 50 means to multiply by the base operator 10, as indicated by the single zero and also by the unit operator 5. The base operation would move the number to be multiplied up 1 level in the table, when the 4 in the 10's level moves to the 4 in the 100's level. The problem in terms of maths is now reduced to '4 times 5'. This gives the value 20 and that 'symbol' should replace the 'symbol' 4 in the current number 400, leading to a result value of 2000.

**4.   Multiplication Example: 2507 x 852 = 2135964**

This is a much more complicated multiplication example that would use different parts of the table and so number parts would need to be indexed as well. The base orders can be recognised by the number of digits in the number and any digit that is not zero has to be separated into a distinct part. Therefore, the first step is to break these two numbers down into:

(2000, 500 and 7) x (800, 50 and 2), leading to the following equation:
(2000 x 800) + (2000 x 50) + (2000 x 2) + (500 x 800) + (500 x 50) + (500 x 2) + (7 x 800) + (7 x 50) + (7 x 2).

Rather like a matrix, all parts of the multiplicand are multiplied by the multiplier. The larger numbers are further broken down into orders of magnitude and the remainders, represented by the non-zero digit and order of magnitude d(m), leading to the following equation:

(2(3) x 8(2)) + (2(3) x 5(1)) + (2(3) x 2) + (5(2) x 8(2)) + (5(2) x 5(1)) + (5(2) x 2) + (7 x 8(2)) + (7 x 5(1)) + (7 x 2).

These could be placed in the appropriate cells and multiplied and added s the algorithm of section 3.4 indicates, but it leads to the following equation:

16(5) + 10(4) + 4(3) + 40(4) + 25(3) + 10(2) + 56(2) + 35(1) + 14.

As all multiplications are at the unit level, they have to be carried out as whole operations, but they then replace the 'symbol' in the number that is the multiplicand. This leads to the following equation:

(16)00000 + (10)0000 + (4)000 + (40)0000 + (25)000 + (10)00 + (56)00 + (35)0 + (14).

This would be broken down into parts again and multiple cells at a level added together, before new additions are repeated. Simply placing these values in the number leads to a final result of 2135964.

**5.   Division Example: 2075 / 25 = 83**

This is a division example of 275 divided by 25. The divisor of 25 cannot be broken down into 20 and 5 and so it is only possible to break the dividend down into 2000, 70 and 5. Then it is a matter of dividing each of these by 25.

- The number 2000 can be moved down a level to 200 and still stay above the divisor value of 25. So the division maths is done over these two numbers, leading to a result of 8(1).
- The number 70 cannot be moved down a level and so it must be divided directly leading to a result of 2 remainder 20.
- The number 5 cannot be divided by 25 and leads to a remainder of 5.

After this phase, there is a result value of 8(1) and 2, plus a remainder value of 20 and 5. Adding the remainders together means that they can maybe be divided again, leading to: 20 + 5 = 25 / 25 = 1 with no remainder. The 2 and the 1 in the Units row would then be added together, to give a total of 3.

This gives the final result parts of 80 and 3, leading to a final result of 83.